\documentclass[
]{ceurart}

\usepackage{listings}
\newcommand{\roberta}[0]{\texttt{RoBERTa}\xspace}

\newcommand{\BERT}[0]{\texttt{BERT}\xspace}

\newcommand{\pe}[1]{\textcolor{black}{#1}}

\begin{document}

\copyrightyear{2024}
\copyrightclause{Copyright for this paper by its authors.
  Use permitted under Creative Commons License Attribution 4.0
  International (CC BY 4.0).}

\conference{CLEF 2024: Conference and Labs of the Evaluation Forum, September 09–12, 2024, Grenoble, France}

\title{HYBRINFOX at CheckThat! 2024 - Task~1: Enhancing Language Models with Structured Information for Check-Worthiness Estimation}


\title[mode=sub]{Notebook for the HYBRINFOX Team at CheckThat! 2024 - Task~1}

\author[1,2]{Géraud Faye}[%
orcid=0000-0002-2985-5964,
email=geraud.faye@centralesupelec.fr
]
\cormark[1]
\address[1]{Airbus Defence and Space, France}
\address[2]{Université Paris-Saclay, CentraleSupélec, MICS, France}

\author[3]{Morgane Casanova}[%
email=morgane.casanova@irisa.fr
]
\address[3]{Université de Rennes, CNRS, Inria, IRISA, France}

\author[4]{Benjamin Icard}[%
orcid=0009-0005-4530-5646,
email=benjamin.icard@lip6.fr
]
\address[4]{LIP6, Sorbonne Université, CNRS, France}
\address[5]{Institut Jean-Nicod, CNRS, ENS-PSL, EHESS, France}

\author[6]{Julien Chanson}[%
email=julien.chanson@mondeca.com
]
\address[6]{Mondeca, France}

\author[1]{Guillaume Gadek}[
email=guillaume.gadek@airbus.com
]

\author[3]{Guillaume Gravier}[
email=guillaume.gravier@irisa.fr
]

\author[5]{Paul \'Egr\'e}[%
orcid=0000-0002-9114-7686,
email=paul.egre@ens.psl.eu
]

\cortext[1]{Corresponding author.}

\begin{abstract}
  This paper summarizes the experiments and results of the HYBRINFOX team for the CheckThat!~2024 - Task~1 competition. We propose an approach enriching Language Models such as RoBERTa with embeddings produced by triples \textit{(subject ; predicate ; object)} extracted from the text sentences. Our analysis of the developmental data shows that this method improves the performance of Language Models alone.  
  On the evaluation data, its best performance was in English, where it achieved an F1 score of 71.1 and ranked 12th out of 27 candidates. On the other languages (Dutch and Arabic), it obtained more mixed results. Future research tracks are identified toward adapting this processing pipeline to more recent Large Language Models. 
\end{abstract}


\maketitle

\section{Introduction}

The recent democratisation of social media \pe{has given the users unprecedented access to information, with the possibility to contribute knowledge as well as to share personal views and opinions. By the same token, however}, it
has also offered misinformation new paths to propagate, sometimes with massive impact. Because of that, automated misinformation detection and automated fact-checking have become tasks of central interest in the data science community. 

This paper deals with a specific aspect of fact-checking, namely with the problem of check-worthiness estimation, presented as Task~1 of the broader CheckThat!~2024 workshop~\cite{clef-checkthat:2024-lncs}, concerned with information quality evaluation. 
Various works dealing with fact-checking operate under the assumption that the entirety of the claims, sentences, or articles in a dataset are \textit{checkworthy}~\cite{10.1007/978-3-030-30796-7_20,kim-choi-2020-unsupervised,thorne-vlachos-2018-automated}. \pe{But this approach can be inefficient, and a useful preliminary step is the identification of which claims are check-worthy, and which are not. The notion of check-worthiness is complex. Some claims in a text are not check-worthy simply because they are not declarative sentences and do not report even potential facts (viz. questions). Others are not check-worthy because, while making declarative assertions, they make claims of no consequence. Conversely, a declarative sentence with potentially harmful consequences is one that ranks high on check-worthiness. Others, finally, may not be check-worthy when they simply report subjective views that are not susceptible of verification proper. It is a non-trivial challenge, therefore, to determine which claims in a document are specifically check-worthy.}



Check-worthiness is a recent task~\cite{10.1145/2806416.2806652}, mostly covered using Language Models~\cite{alam-etal-2021-fighting-covid}. In this paper, we propose an approach designed to leverage structured information from the text, in order to enhance the representation obtained with a Language Model. Because the task of check-worthiness estimation is related to fact-checking, it seems appropriate to identify facts from the text to help the model predict check-worthiness. By using both structured facts and Language Models embeddings, we obtained better results than when using Language Models alone, ranking 12th among 27 competing teams for English (with an F1 score of 71.1). Results were more mixed for the non-English languages represented in the test set: \pe{in Dutch our method ranked 8th out of 16 candidates (F1 score of 58.9), and in Arabic it ranked 10th out of 14 (F1 score of 51.9).}


In Section~\ref{sec:related_works}, we open with a quick review of the state of the art on the task. In Section~\ref{sec:methodo}, we spell out the functioning of the proposed processing pipeline. Then, Section~\ref{sec:results} discusses preliminary results obtained with the initial training data and presents the final submitted results. Finally, some elements on the evolution and future use of the proposed hybrid system are presented in Section~\ref{sec:future}.

\section{Related work}
\label{sec:related_works}

As explained in the previous section, check-worthiness is a fairly recent task, first mentioned in 2015~\cite{10.1145/2806416.2806652}. Several datasets have been constructed, like the ClaimBuster dataset~\cite{Arslan2020ABD}, or the datasets proposed at the CheckThat workshops since 2018~\cite{clef2018}. These datasets focus on two main types of \pe{contexts for the task}:


\begin{itemize}
    \item Classifying sentences from a political debate. These could be used to ease fact-checking during television political debates, on datasets such as~\cite{rayar_tv_political}.
    \item Classifying tweets. Because they are easily and widely shared online, check-worthiness is an important task for online discussions to avoid information manipulation.
\end{itemize}

Both of these categories are important, and a commonality between them is their short format. 
However, the scope of this task of check-worthiness can be widened so as to also include online press, with an eye to so-called ``pink slime'' news~\cite{Horne2024NELAPSAD}, encompassing longer texts whose truthfulness is questionable.

Among pioneering approaches to the task, we find methods such as ClaimRank~\cite{jaradat-etal-2018-claimrank}, using traditional NLP methods (e.g. lemmatization, TF-IDF) to identify check-worthy claims.

More recent approaches take advantage of the Transformer layer and of pretrained Language Models such as BERT~\cite{devlin_bert:_2018}, RoBERTa~\cite{roberta} or XLNet~\cite{xlnet}. The fortune of these approaches can be seen in the 2023 CheckThat! Task 1 overview paper~\cite{Alam2023OverviewOT}, showing that nearly all teams used a transformer-based Language Model.

With the even more recent development of Large Language Models and of Generative AI, a natural shift has been made toward the use of LLMs, relying on prompt engineering~\cite{10.1145/3560815} and in-context learning~\cite{Dong2023ASF} to achieve check-worthiness estimation. These approaches were used by the winners of this year's competition, both in English~\cite{clef-checkthat:2024:task1:factfinders} and in Dutch~\cite{clef-checkthat:2024:task1:turquaz}.

\section{Methodology}
\label{sec:methodo}

\subsection{Model}

\begin{figure}
    \centering
    \includegraphics[width=\linewidth]{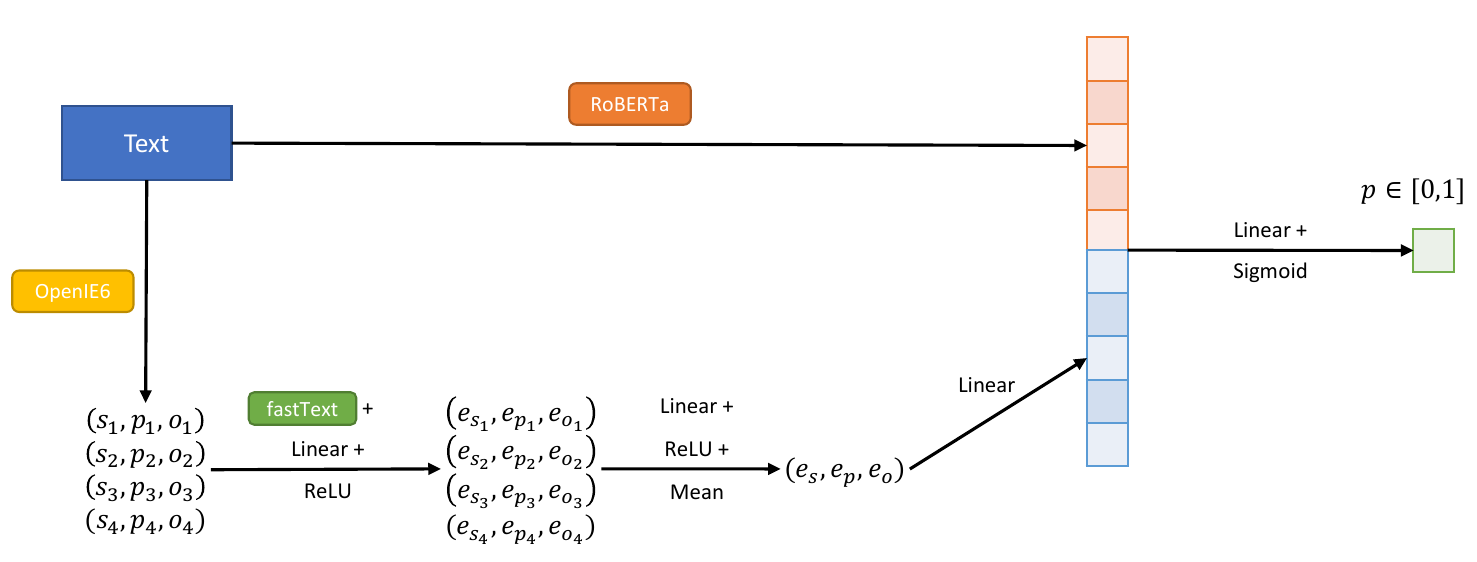}
    \caption{Our proposed architecture: adding structured information extracted from the text to enhance the LM embeddings. RoBERTa and OpenIE6 can be switched with other models for non-English languages.}
    \label{fig:archi}
\end{figure}

A straightforward approach for check-worthiness estimation would be to use a pretrained Language Model fine-tuned with the provided training data. However, these language models produce embeddings that are opaque, even if they are sufficient most of the time. To increase the quality of the language model predictions, we propose to use them in conjunction with a small neural network able to leverage structured information from the input text. A visual description of the architecture is given in Figure~\ref{fig:archi}. The processing pipeline is the following:

\begin{enumerate}
    \item To begin with, the text is embedded using a Language Model. In our case, the \roberta model~\cite{roberta} is used, producing an embedding of dimension 768. \roberta was chosen for its ease of use, its high performance for classification tasks and its relatively small size when compared to recent LLMs.
    \item In parallel, the text is structured using an Open Information Extraction system. These systems extract information from the text in the form of triples (\textit{subject}; \textit{predicate}; \textit{object}). We used OpenIE6~\cite{kolluru&al20} to extract triples in the English language. Using triples allows us to produce structured information from the text and to reduce syntactic complexity, with the aim of helping sentence classification. A maximum limit of 4 triples by text were extracted, which is enough to consider all information triples for more than 90\% of sentences. Each part of the triples is encoded using fastText~\cite{fasttext}, producing 3 vectors of dimension 300 per triple. These vector representations go through a dense layer with ReLU activation function. Then,  they are averaged before being combined in the last layer to produce an embedding of dimension 768 (the same dimension as the Language Model).
    \item Encodings from the previous two parts are concatenated and go through a dense layer with ReLU activation function with a final output producing the probability of being checkworthy \pe{by means of a sigmoid activation function}.
\end{enumerate}

The described architecture can be transposed to other languages when an OpenIE system and an LM are available. In the context of Task 1 of the evaluation, for Spanish, Dutch and Arabic, \roberta was swapped with a multilingual \BERT~\cite{devlin_bert:_2018}
.\footnote{\url{https://huggingface.co/google-bert/bert-base-multilingual-cased}} The OpenIE6 system was replaced with Multi²OIE \cite{ro-etal-2020-multi} in a zero-shot setting for non-English languages, providing worse performance than OpenIE6 on English, but allowing us to test the architecture on other languages. \pe{In principle, the same architecture can be used for any language, and in practice it is applicable to the 98 languages currently supported by the multilingual version of BERT.}


\subsection{Example}

To better understand how this architecture works, we illustrate the pipeline with a simple example. We took a sentence from the training English dataset: \textit{"I must remind him the Democrats have controlled the Congress for the last twenty-two years and they wrote all the tax bills."} This sentence comes from a debate between US presidential candidates Jimmy Carter and Gerald Ford on September 26$^{th}$ 1976. It is deemed to be checkworthy, as it contains allegations on Jimmy Carter's party.

The classical NLP classification pipeline using RoBERTa would consist in producing a 768 dimension embeddings and doing classification passing this embeddings through a neural network. In our approach, we produce in parallel embeddings of the triples that can be extracted from the text. For this example, OpenIE6 produces the three following triples:
$$\hbox{\textit{(I; must remind; him the Democrats have controlled the Congress for the last twenty-two years)}}$$
$$\hbox{\textit{(the Democrats; have controlled; the Congress for the last twenty-two years)}}$$
$$\hbox{\textit{(they; wrote; all the tax bills)}}$$

Only the last two triples contain the information that is check-worthy. However, there are no triple-level annotations, so we keep all triples. Firstly, we encode each subject, predicate and object with fastText, creating in this case three embedding vectors for each triple. These embeddings go through the same linear layer and the embeddings of subjects, predicates and objects are averaged by component (subject, predicate, object).
This leads to three vectors of dimension 300, representing the subjects, predicates and objects of the triples extracted from the text. Finally they are concatenated and projected into a vector of dimension 768, the same dimension as the \roberta embeddings.
\roberta embeddings and embeddings for triples are eventually concatenated for standard classification.

\section{Results}
\label{sec:results}

This section is divided in three parts. The first presents the 
protocol of our model. The second part reports the evaluation of our proposed approach and of a standard Language Model, in order to measure how the additional triple processing part impacts performance. The \pe{third} part contains an analysis of our submitted results, as well as of the difficulties encountered.

\subsection{Training procedure}

Each model was trained over 5 epochs on the train set. After each epoch, the model was evaluated on the dev set and the best model in terms of macro-F1 score was kept. The scores reported in Section~\ref{sec:early_results} are the macro-F1 score on the dev-test set.

In our procedures, only the train set was used for training the model, the dev and dev-test sets being used for model selection. Reported results were produced by the models with the best dev-test macro-F1 score, which were also used to make predictions on the final test set.

\subsection{Preliminary results on the development data}
\label{sec:early_results}

The main goal of our approach was to evaluate how combining structured information from the text with a standard Language Model is impacting performance. Results observed on the dev-test set provided before the competition are provided in Table~\ref{tab:early_results}. 

\begin{table*}[h]
\caption{Macro-F1 scores on the dev-test split with a Language Model (\texttt{RoBERTa}$^{(1)}$ or bert-base-multilingual-cased$^{(2)}$, where the superscripts indicate which LM was used for which language) and our own approach (LM+Triples). The F1-scores are multiplied by 100 to homogenize with the organizers' way of communicating scores.}
  \label{tab:early_results}
  \begin{tabular}{c|c|c|c|c}
         & English$^{(1)}$ & Arabic$^{(2)}$ & Dutch$^{(2)}$ & Spanish$^{(2)}$ \\
         \hline
         LM & 84.042 & 58.273 & \textbf{40.866} & 59.975 \\
         \hline
        LM+Triples & \textbf{86.458} & \textbf{62.300} & 39.832 & \textbf{62.371} \\
         \hline
         Performance gain & \textbf{+2.416} & \textbf{+4.027} & -1.034 & \textbf{+2.396}
    \end{tabular}
\end{table*}

In most cases, our approach outperformed the LM baseline, achieving the highest performance gain in Arabic, followed by English and Spanish.

It appears that performance is generally lower for non-English languages. This is due to the fact that multilingual models perform worse, but they allowed us to process different languages with the same architecture and weights.

The same goes for the OpenIE system, with OpenIE6 being specifically trained on English, and Multi²OIE being used in a zero-shot setting (no non-English training sample was used for training). This limitation is pointed out in the Multi²OIE paper~\cite{ro-etal-2020-multi}, but it is the only existing open-source OpenIE system able to process Arabic, Dutch, English and Spanish. As it relies on a multilingual \texttt{BERT}, it also suffers from lower performance for relatively low-resource languages, explaining the decrease in performance for Dutch.

\subsection{Results on the evaluation data}

The competition scores and ranking are shown in Table~\ref{tab:results}, with the scores of the best performing team (state-of-the-art for this dataset) and the baseline being also reported.

\begin{table*}[h]
    \centering
\caption{Final results on the evaluation set. The reported F1-scores and ranking were provided by the organizers of the task.}
    \label{tab:results}
    \begin{tabular}{c|c|c|c}
        Language & English & Arabic & Dutch \\
        \hline
        Best performance & 80.2 (FactFinders~\cite{clef-checkthat:2024:task1:factfinders}) & 56.9 (visty) & 73.2 (TurQUaz~\cite{clef-checkthat:2024:task1:turquaz})\\
        \hline
        Hybrinfox & 71.1 (12/27) & 51.9 (10/14) & 58.9 (8/16)\\
        \hline
        Baseline & 30.7 (27/27) & 41.8 (13/14) & 43.8 (14/16)
    \end{tabular}
\end{table*}

For the three languages, our approach outperformed the baseline by a substantial margin. Performance for non-English languages was mixed. The Arabic dataset proved to be challenging for all teams, with most candidate approaches getting scores between 50 and 55.

\subsection{Discussion}

While the proposed approach outperformed \roberta on the dev-test set, several upgrades could have been made to reduce possible errors in the processing pipeline. Firstly, it is well known that triples extracted with OpenIE are noisy and may not always contain useful facts for the task. This can be seen in the first triple of the example given in Section~\ref{sec:methodo}: \textit{(I; must remind; him the Democrats have controlled the Congress for the last twenty-two years)}. One way would be to filter out the triples that do not contain named entities in the subject and object part. This approach would keep only the second triples in the example. One additional step to increase the usefulness of triples would be to apply a coreference analysis, changing pronouns by the objects they refer to. After coreference, \textit{(they; wrote; all the tax bills)} would become \textit{(the Democrats; wrote; all the tax bills)}, which is more descriptive.

Another way of improving this approach would be to use post-hoc explanation methods such as integrated gradients~\cite{9378686}, to identify which embeddings make the highest contribution to the prediction. 
This could help identify the triples most relevant to the prediction, giving interpretability to the proposed addition, and further input for a fact-checking system.

\section{Future work and conclusion}
\label{sec:future}

The HYBRINFOX team is interested in neurosymbolic architectures and our aim is generally to improve performance of Language Models by adding structured information from the texts. This approach has to be adapted to misinformation detection or fact-checking settings. In general, we believe that all tasks that are related to factual claims could benefit from adding structured information into their pipeline in order to increase performance.

The proposed approach uses Language Models such as \BERT (for OpenIE) or \roberta, but could be upgraded by using most recent advances in Large Language Models such as Mistral or ChatGPT. An LLM prompted with instructions could easily perform a similar pipeline:

\begin{enumerate}
    \item Extract information triples from the text.
    \item Select factual triples.
    \item Identify if the factual triples are check-worthy.
\end{enumerate}

This approach could help identify which part of the text contains check-worthy information with better accuracy. 


To conclude, the proposed approach, enriching Language Models with a level of structured information, has shown promising results in comparison to the use of Language Models alone on the task of check-worthiness estimation. 
For check-worthiness, the extraction of factual triples from the text helps classification. However, performance was mixed on non-English texts. 

Further analyses need to be conducted with other expert systems to further improve performance. As mentioned in the introduction, the definition of what counts as check-worthy is complex. One approach, which we have not tried, might be to consider as checkworthy first and foremost sentences making objective claims. For that purpose, we may piggyback on the methods used in Task 2 of the CheckThat! Lab~\cite{clef-checkthat:2024:task2} dealing with the classification of subjective vs objective sentences. We leave that exploration for further work.


\section*{Acknowledgements}

\pe{We thank two anonymous referees for helpful comments and suggestions on the first version of this report.} This work was supported by the programs HYBRINFOX (ANR-21-ASIA-0003), FRONTCOG (ANR-17-EURE-0017), and THEMIS (n°DOS\-0222794/00 and n° DO\-S0\-222795/00). PE thanks Monash University for hosting him during the writing of this paper, in the context of the program {PLEXUS (Marie Sk\l odowska-Curie Action, Horizon Europe Research and Innovation Programme, grant n°10\-1086295).

\bibliography{biblio}

\end{document}
